\documentclass[10pt,twocolumn,letterpaper]{article}

\usepackage[pagenumbers]{cvpr} 
\usepackage{pifont}

\usepackage{booktabs}       
\usepackage{multirow}       
\usepackage{makecell}       
\usepackage{array}          

\definecolor{cvprblue}{rgb}{0.21,0.49,0.74}
\usepackage[pagebackref,breaklinks,colorlinks,allcolors=cvprblue]{hyperref}

\title{Fairness in Multi-modal Medical Diagnosis with Demonstration Selection}

\author{
Dawei Li\textsuperscript{1}\thanks{These authors contributed equally to this work.} \quad
Zijian Gu\textsuperscript{2}\footnotemark[1] \quad 
Peng Wang\textsuperscript{3}\thanks{These authors also contributed equally to this work.} \quad  
Chuhan Song\textsuperscript{4}\footnotemark[2] \quad  
Zhen Tan\textsuperscript{1} \\ 
Mohan Zhang\textsuperscript{5} \quad 
Tianlong Chen\textsuperscript{5} \quad 
Yu Tian\textsuperscript{6} \quad 
Song Wang\textsuperscript{6} \\
\vspace{0.5em}
\textsuperscript{1}Arizona State University \quad
\textsuperscript{2}University of Rochester \quad
\textsuperscript{3}University of Virginia \quad
\textsuperscript{4}UCL \\
\textsuperscript{5}University of North Carolina at Chapel Hill \quad
\textsuperscript{6}University of Central Florida
}

\begin{document}
\maketitle

\begin{abstract}
    Multimodal large language models (MLLMs) have shown strong potential for medical image reasoning, yet fairness across demographic groups remains a major concern. Existing debiasing methods often rely on large labeled datasets or fine-tuning, which are impractical for foundation-scale models. We explore In-Context Learning (ICL) as a lightweight, tuning-free alternative for improving fairness. Through systematic analysis, we find that conventional demonstration selection (DS) strategies fail to ensure fairness due to demographic imbalance in selected exemplars. To address this, we propose Fairness-Aware Demonstration Selection (FADS), which builds demographically balanced and semantically relevant demonstrations via clustering-based sampling. Experiments on multiple medical imaging benchmarks show that FADS consistently reduces gender-, race-, and ethnicity-related disparities while maintaining strong accuracy, offering an efficient and scalable path toward fair medical image reasoning. These results highlight the potential of fairness-aware in-context learning as a scalable and data-efficient solution for equitable medical image reasoning.
\end{abstract}

\section{Introduction}
\label{sec:intro}

Recent advances in medical image reasoning have been driven by the rapid development of \textit{multimodal large language models (MLLMs)}~\cite{li2023llava,wu2023multimodal,zhang2024mm,yin2024survey}, which unify visual perception and linguistic reasoning within a single foundation architecture. By leveraging large-scale vision–language pretraining and instruction tuning, these models demonstrate impressive capabilities in medical report generation, visual question answering, and disease localization. The ability of MLLMs to generalize across modalities and clinical tasks positions them as powerful tools for improving diagnostic accuracy, efficiency, and accessibility in medical imaging workflows~\cite{liu2023medical,mesko2023impact,xiao2024comprehensive}.

However, fairness and equity remain fundamental challenges in deploying medical AI responsibly. Medical imaging data inherently reflect population imbalance, device heterogeneity, and institutional biases, which can lead to uneven model performance across demographic subgroups. Prior studies~\cite{drukker2023toward,koccak2025bias} have documented that such disparities often disadvantage underrepresented populations, resulting in unequal diagnostic outcomes and reduced trust in AI-assisted healthcare systems. Addressing fairness in MLLMs is particularly important because biased predictions in high-stakes clinical contexts can propagate or even amplify existing healthcare inequities.

Existing fairness-improvement techniques—such as reweighting~\cite{kamiran2012data}, adversarial debiasing~\cite{madras2018learning}, and domain adaptation~\cite{ganin2016domain}—face major limitations when applied to MLLMs. These methods typically depend on large-scale labeled data and costly retraining, which are impractical in medical imaging due to the scarcity of expert annotations. Moreover, foundation-scale MLLMs are often black-box systems with inaccessible internal parameters, making direct fairness regularization infeasible. Even when fine-tuning is technically possible, it may lead to catastrophic forgetting~\cite{huang2024mitigating}, erasing the broad multimodal reasoning ability crucial for other downstream tasks. These challenges highlight the need for lightweight, data-efficient, and tuning-free strategies to improve fairness in medical image reasoning.

\begin{figure*}[t]
    \centering
    \includegraphics[width=1.0\linewidth]{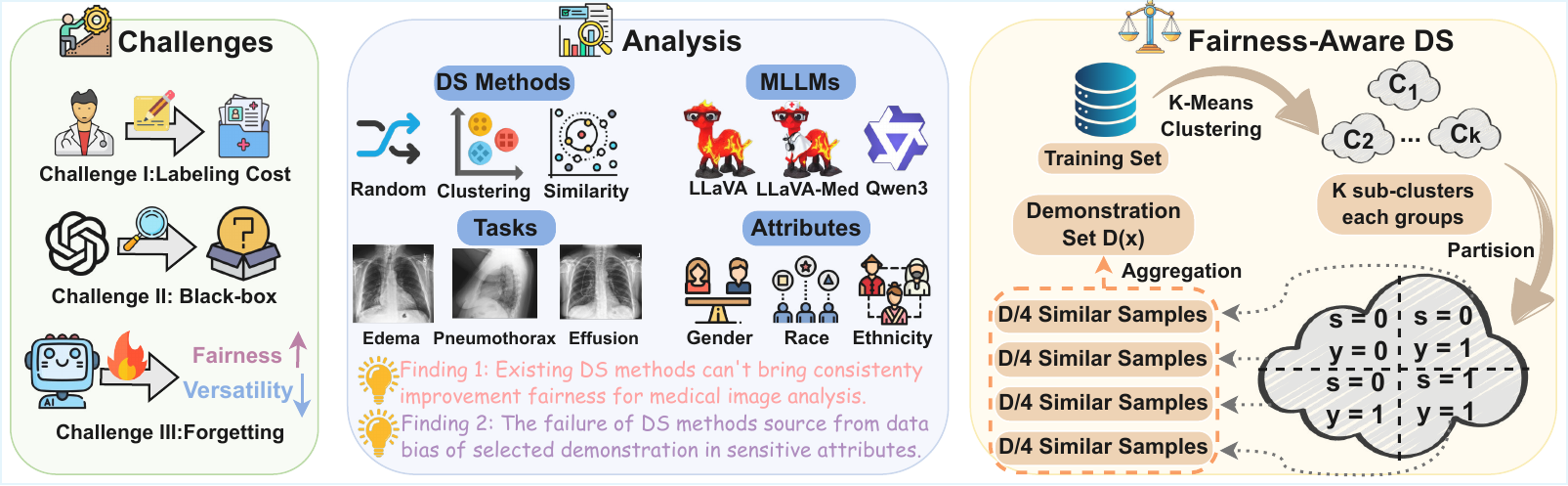}
    \caption{\textbf{Overview of challenges and findings.} We investigate whether ICL can improve fairness in medical image reasoning. Different demonstration selection (DS) strategies—Random, Similarity, and K-Means—are analyzed across attributes (Gender, Race, Ethnicity). Empirical results reveal that data imbalance in selected demonstrations is the primary source of fairness degradation.}
    \label{fig:icl_fairness}
\end{figure*}

In this work, we explore \textit{In-Context Learning (ICL)}~\cite{min2022metaicl,wies2023learnability,dong2024survey} as a promising and practical mechanism to enhance fairness without model retraining. ICL enables adaptation through contextual exemplars, or demonstrations, presented at inference time, allowing MLLMs to adjust behavior flexibly based on provided examples. We first conduct comprehensive analyses to examine how different demonstration selection (DS) strategies~\cite{zhang2023makes}—such as random, similarity-based, and clustering-based methods—affect fairness across multiple medical imaging datasets and demographic attributes. Our findings reveal that existing DS heuristics cannot guarantee fairness consistency, as demographic imbalances within selected demonstrations propagate bias to model predictions. Motivated by these insights, we design a new method, \textit{Fairness-Aware Demonstration Selection (FADS)}, which constructs balanced and semantically relevant exemplars through demographic-aware clustering and subgroup-level sampling. Extensive experiments show that FADS effectively mitigates both data-driven and model-induced bias, providing a stable fairness–performance trade-off. We further analyze the mechanisms behind FADS, offering empirical insights into how balanced demonstrations contribute to equitable multimodal reasoning.

Our main contributions are summarized as follows:
\begin{itemize}
    \item We identify and formalize fairness challenges unique to applying MLLMs in medical image reasoning, emphasizing the limitations of conventional resource-intensive fairness interventions.
    \item We introduce ICL as a lightweight, tuning-free strategy for fairness enhancement and systematically study the impact of demonstration selection on bias propagation.
    \item We propose a fairness-aware ICL framework (FADS) that mitigates both data-driven and model-induced biases through balanced exemplar construction, achieving improved fairness and stability across tasks and datasets.
\end{itemize}

\section{Analysis: Can In-Context Learning Boost Fairness in Medical Image Analysis?}

In this section, we empirically examine whether ICL can improve fairness in MLLMs for medical image reasoning. Specifically, we analyze how different DS strategies influence fairness and task performance across demographic groups. We begin by introducing three commonly used DS strategies and then present quantitative analyses that reveal their limitations and underlying causes of unfairness.

\subsection{Demonstration Selection}
\label{sec:demo_selection}

Given a labeled pool $\mathcal{X}_L = \{(x_i, y_i, s_i)\}_{i=1}^{N}$, where $x_i$ denotes the textual or multimodal input, $y_i$ is the task label, and $s_i$ represents a sensitive attribute, the goal of demonstration selection is to construct, for each test query $x$, a demonstration set $\mathcal{D}(x) = \{(x_j, y_j, s_j)\}_{j=1}^{K}$ of size $K$ to be used in ICL. We investigate three representative strategies below.

\noindent\textbf{Random Selection.}~\cite{wei2022chain}
The most straightforward strategy randomly samples $K$ examples from the labeled pool without considering semantic or demographic information:
\[
\mathcal{D}_{\text{rand}}(x) = \text{Sample}\big(\mathcal{X}_L, K\big),
\]
where $\text{Sample}(\cdot)$ denotes uniform sampling without replacement. This method is computationally efficient and widely used as a baseline. However, its demonstrations may be semantically irrelevant to the query, introducing noise and potentially amplifying demographic imbalance.

\noindent\textbf{Similarity-based Selection.}~\cite{luo2024context}
A more targeted strategy selects demonstrations that are semantically close to the query $x$. Each $x_i \in \mathcal{X}_L$ is encoded as an embedding $\mathbf{e}_i = M_{\text{enc}}(x_i)$ using a pretrained text or multimodal encoder $M_{\text{enc}}(\cdot)$. The cosine similarity between the query and each candidate is computed as
\[
f(x, x_i) = \frac{\mathbf{e}_x \cdot \mathbf{e}_i}{\|\mathbf{e}_x\|\|\mathbf{e}_i\|}.
\]
The $K$ most similar samples are then selected:
\[
\mathcal{D}_{\text{sim}}(x) = 
\arg\max_{\mathcal{D} \subseteq \mathcal{X}_L, |\mathcal{D}|=K}
\sum_{(x_i, y_i, s_i) \in \mathcal{D}} f(x, x_i).
\]
This approach prioritizes contextual relevance and is commonly adopted in retrieval-based ICL frameworks. Yet, as we later show, focusing solely on semantic similarity can inadvertently favor majority subgroups prevalent in the dataset.

\begin{figure*}[t]
    \centering
    \includegraphics[width=1.0\linewidth]{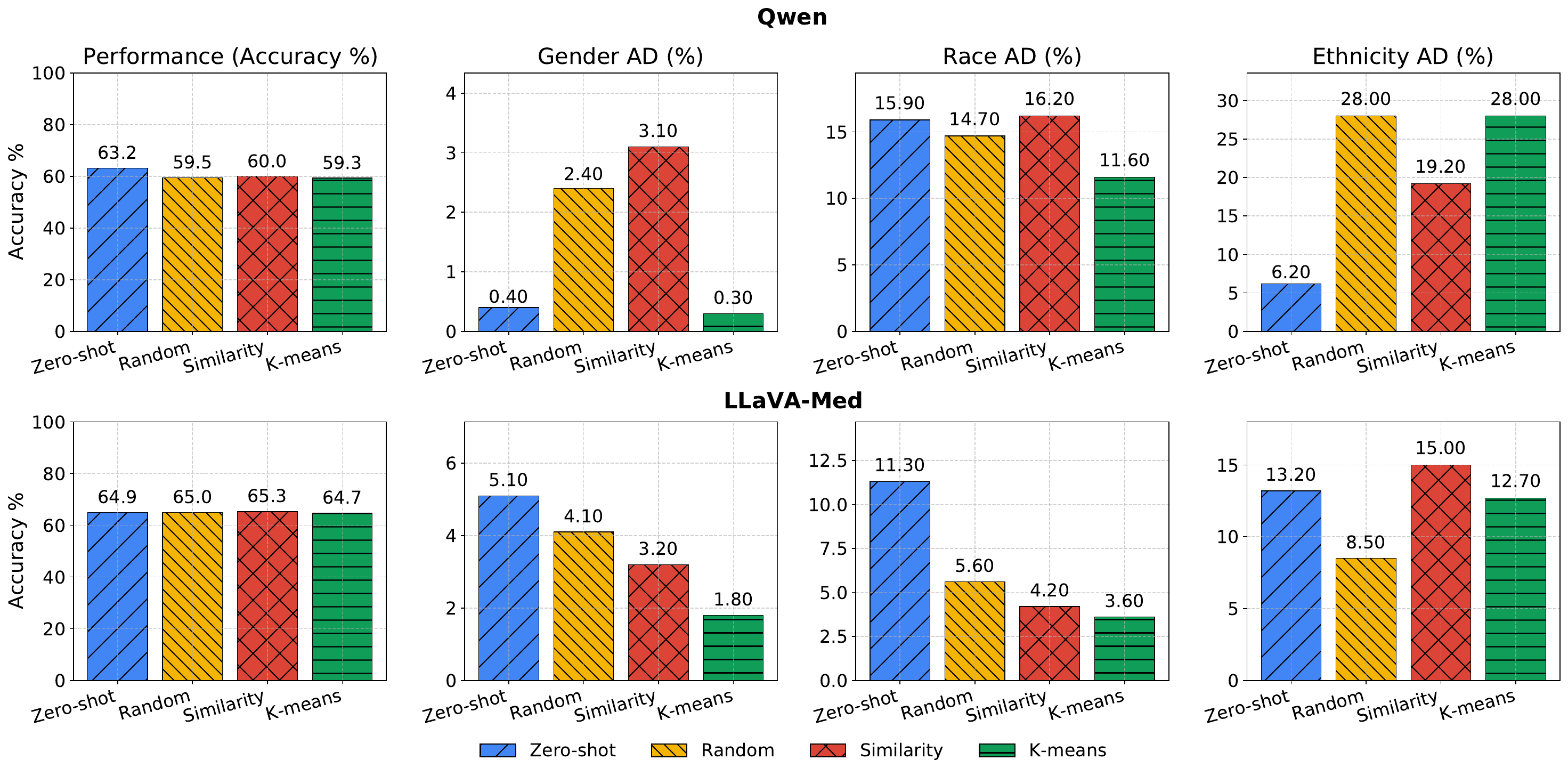}
    \caption{\textbf{Comparison of performance and fairness metrics} for Random, Similarity, and K-Means demonstration selection methods on Qwen (bottom row) and LLaVA-Med (top row). Conventional DS strategies exhibit inconsistent and sometimes conflicting fairness behaviors.}
    \label{fig:finding1}
\end{figure*}

\noindent\textbf{Clustering-based Selection.}~\cite{zhang2022automatic}
To ensure greater diversity, clustering-based strategies partition the candidate pool into $C$ clusters $\{C_1, C_2, \ldots, C_C\}$ in the embedding space (e.g., via K-Means) and sample a fixed number of demonstrations from each cluster. Let $K_c = \lfloor K / C \rfloor$ denote the number of demonstrations per cluster:
\[
\mathcal{D}_{\text{clust}}(x) = 
\bigcup_{c=1}^{C} \text{Sample}\big(C_c, K_c\big).
\]
Each $C_c$ contains semantically similar samples within the cluster but diverse examples across clusters, ensuring broader coverage of both semantic and demographic subgroups.

\subsection{Analysis Results}

\begin{figure}[t]
    \centering
    \includegraphics[width=1.0\linewidth]{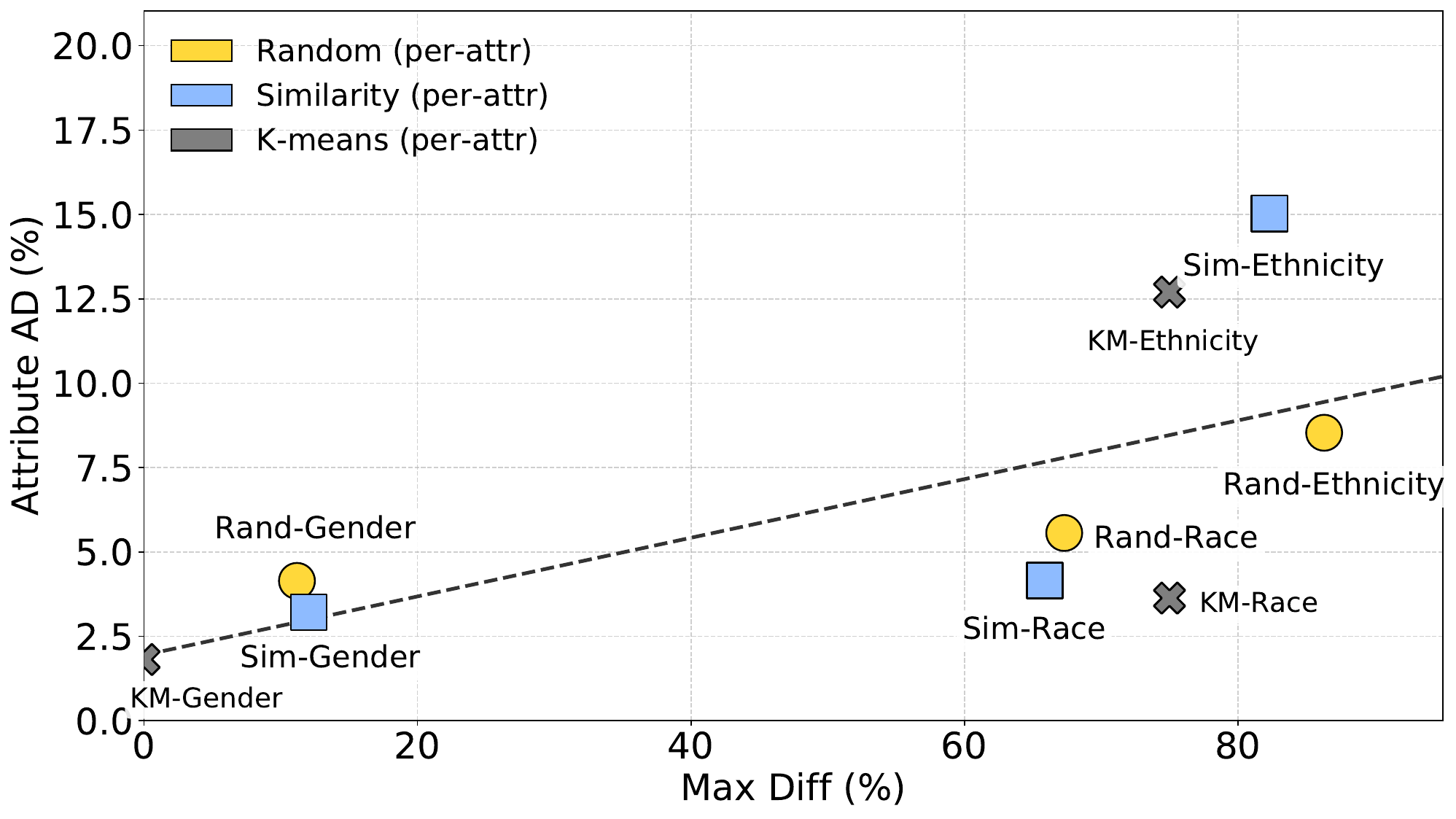}
    \caption{\textbf{Correlation between data imbalance and fairness disparity.} Each point corresponds to one attribute under a DS method (Random, Similarity, or K-Means). Larger demographic imbalance (MaxDiff) in the selected demonstrations correlates with higher Average Disparity (AD), indicating that biased exemplar composition directly drives fairness degradation.}
    \label{fig:finding2}
\end{figure}

\noindent\textbf{Experimental Setup.}
We evaluate whether ICL can enhance fairness under different DS strategies using two multimodal large language models: \textbf{Qwen2.5-VL-7B} and \textbf{LLaVA-Med}. Each method constructs 8–16 demonstrations per query without fine-tuning. Fairness is measured by \textbf{Average Disparity (AD)} across \textbf{Gender}, \textbf{Race}, and \textbf{Ethnicity}, and performance is evaluated using \textbf{Accuracy}, \textbf{Precision}, \textbf{Recall}, and \textbf{F1-score}. Figure~\ref{fig:finding1} presents quantitative comparisons, while Figure~\ref{fig:finding2} illustrates the relationship between data bias and fairness.

\noindent\textbf{Finding 1: Existing DS methods fail to consistently improve fairness.}
Across both models, DS strategies display unstable and often contradictory fairness trends (Figure~\ref{fig:finding1}). For \textbf{Qwen}, the Similarity-based method achieves the highest accuracy (65.3\%) but also the largest Ethnicity AD (15.0\%), indicating that stronger performance does not guarantee equitable outcomes. K-Means reduces Gender and Race ADs but lowers overall accuracy, while Random selection shows mixed effects across attributes. A similar pattern emerges in \textbf{LLaVA-Med}, where Random achieves smaller Race AD (14.7\%) but extremely high Ethnicity AD (28.0\%). These findings demonstrate that conventional DS heuristics cannot ensure fairness stability, as outcomes depend more on dataset composition and model sensitivity than on true fairness enhancement.

\noindent\textbf{Finding 2: Unfairness arises from demographic imbalance in selected demonstrations.}
To quantify demographic imbalance, we compute the \textbf{Max Diff} metric, defined as the difference between the maximum and minimum proportional representation among sensitive groups:
\[
\text{MaxDiff} = \max(\text{ratio}) - \min(\text{ratio}).
\]
A smaller value indicates balanced demographics, while a larger value signifies stronger bias. For example, a 75/25 male–female split yields $\text{MaxDiff} = 0.50$, whereas perfect parity corresponds to $\text{MaxDiff}=0$.

Figure~\ref{fig:finding2} visualizes the relationship between \textit{data bias} (MaxDiff) and \textit{fairness disparity} (AD). A strong positive correlation ($R=0.67$) emerges: higher demographic imbalance leads to greater fairness gaps. For instance, Random-Ethnicity (MaxDiff=86.3\%) and Similarity-Ethnicity (82.3\%) yield the largest ADs (8.5–15.0\%), while more balanced attributes such as Gender exhibit smaller disparities. These results reveal that DS methods, despite improving contextual relevance, often propagate demographic skew from exemplars to model predictions.

\noindent\textbf{Summary.}
Our analysis demonstrates that fairness inconsistency in current DS strategies stems from demographic imbalance in the selected demonstrations. Data bias directly correlates with fairness degradation, motivating the need for a fairness-aware selection mechanism. This insight forms the basis of our proposed \textbf{\underline{F}airness-\underline{A}ware \underline{D}emonstration \underline{S}election} framework, introduced in the next section.

\section{Fairness-Aware Demonstration Selection}
\label{sec:method}

Building upon our analysis in Section~\ref{sec:demo_selection}, which revealed that demographic imbalance in selected exemplars is the primary source of fairness degradation, we propose the \textbf{\underline{F}airness-\underline{A}ware \underline{D}emonstration \underline{S}election (FADS)} framework. FADS is a lightweight, tuning-free method designed to enhance fairness in MLLMs through ICL. Rather than retraining or fine-tuning the model, FADS strategically selects balanced and representative demonstrations to mitigate both data-driven and model-induced biases.

Given a labeled dataset 
$\mathcal{X}_L = \{(x_i, y_i, s_i)\}_{i=1}^{N}$, 
where $x_i$ denotes the input (textual or multimodal), 
$y_i \in \{0,1\}$ is the task label, 
and $s_i \in \{0,1\}$ represents a sensitive attribute (e.g., gender or race), 
the objective of FADS is to construct, for each query $x$, a fairness-aware demonstration set $\mathcal{D}(x)$ that is both semantically relevant and demographically balanced.

\paragraph{Step 1: Data Bias Mitigation.}
To decorrelate sensitive attributes from outcome labels, FADS first clusters the labeled samples into $K$ groups in the embedding space using a pretrained encoder (e.g., Sentence-BERT). Each cluster $C_i$ is then subdivided into four sub-clusters $C_i^{(s,y)}$, corresponding to each $(s,y)$ pair. The goal is to identify clusters with approximately uniform subgroup distributions. FADS selects $N_d$ clusters by minimizing deviation in subgroup proportions:
\[
G = \arg\min_{G \subseteq \{C_i\}} 
\sum_{C_i \in G} \sum_{s,y} 
\frac{1}{|C_i|} \Big|\, |C_i^{(s,y)}| - \tfrac{1}{4}|C_i| \,\Big|.
\]
The resulting subset $G = \{G_1, G_2, \ldots, G_{N_d}\}$ represents a candidate pool where sensitive attributes and outcome labels are more evenly distributed. This step mitigates data bias in the demonstration space before exemplar selection.

\paragraph{Step 2: Balanced Demonstration Selection.}
For each query sample $x$, FADS constructs a balanced demonstration set $\mathcal{D}(x)$ of size $D$. Within each $(s,y)$ subgroup, $M = D / 4$ examples are selected from the filtered clusters based on their cosine similarity to the query in the embedding space:
\[
\mathcal{D}_{s,y}(x) = 
\arg\max_{\mathcal{D}_{s,y}} 
\sum_{c \in G_{s,y}} f(x, c), 
\quad 
\mathcal{D}(x) = \bigcup_{s,y} \mathcal{D}_{s,y}(x),
\]
where $f(x, c)$ measures the similarity between the query and candidate. This ensures that the selected demonstrations are both contextually relevant and demographically balanced, preventing bias propagation from imbalanced exemplars.

FADS systematically addresses the fairness limitations observed in conventional demonstration selection. By incorporating demographic balance into exemplar retrieval, it reduces data-induced unfairness while preserving contextual alignment. Importantly, FADS achieves these improvements without modifying model parameters or requiring additional training, offering a practical plug-and-play approach to enhance fairness in medical image reasoning with MLLMs.

\section{Experiments}
\label{sec:experiments}

\subsection{Experimental Setup}

\begin{table*}[t]
  \centering
  \caption{Main experiment: Qwen on Glaucoma (8-shot). FADS yields the best overall fairness--performance trade-off. All AD values are reported in percentage points (\%).}
  \label{tab:glaucoma_qwen}
  \begin{tabular}{lccccc}
    \toprule
    Method & Accuracy (\%) & Gender AD (\%) $\downarrow$ & Race AD (\%) $\downarrow$ & Ethnicity AD (\%) $\downarrow$ & Avg AD (\%) $\downarrow$ \\
    \midrule
    Zero-shot & 64.9 & 5.14 & 11.33 & 13.18 & 9.88 \\
    Random & 65.0 & 4.14 & 5.56 & 8.53 & 6.08 \\
    Similarity & 65.3 & 3.21 & 4.15 & 15.03 & 7.46 \\
    K-means & 64.7 & \textbf{1.81} & \textbf{3.63} & 12.70 & 6.05 \\
    \textbf{FADS (ours)} & \textbf{66.4} & 3.73 & 7.09 & \textbf{6.42} & \textbf{5.75} \\
    \bottomrule
  \end{tabular}
\end{table*}

\paragraph{Datasets.}
We evaluate our method on the following two widely used medical vision-language datasets.

\begin{itemize}[leftmargin=*]
    \item \textbf{FairCLIP Glaucoma}~\cite{luo2024fairclip}: 10,000 fundus (SLO) images with clinical notes from the Harvard Glaucoma Fairness dataset (7K train / 2K val / 1K test). Each sample is annotated with gender (Male/Female), race (White/Black/Asian), and ethnicity (Hispanic/Non-Hispanic/Unknown).
    
    \item \textbf{CheXpert Plus}~\cite{irvin2019chexpert}: We curate balanced subsets focusing on edema, pleural effusion, and pneumothorax classification with 2,500 samples (1,250 positive, 1,250 negative) per disease.
\end{itemize}

\paragraph{Models.}
We evaluate on three vision-language models: \textbf{Qwen2.5-VL-7B-Instruct}~\cite{qwen2vl}, \textbf{LLaVA-Med}~\cite{li2023llava} (medical domain-adapted), and \textbf{LLaVA-v1.6-Vicuna-7B}~\cite{liu2023llava} (general-purpose). For embedding computation, we use \textbf{GME-Qwen2-VL-2B}~\cite{zhang2024gme}.

\paragraph{Baselines.}
We compare six 8-shot demonstration selection strategies:

\begin{itemize}[leftmargin=*]
    \item \textbf{Zero-shot}: No demonstrations.
    \item \textbf{Random}: Randomly select 8 demonstrations.
    \item \textbf{Similarity}: Select 8 most similar demonstrations based on embedding distance.
    \item \textbf{K-means}: Cluster training data into $K=64$ clusters and select representatives from 8 random clusters.
    \item \textbf{FADS (Ours)}: Fairness-aware selection with $K=64$ balanced clusters, stratified sampling on one sensitive attribute (e.g., gender), and balanced labels (4 positive, 4 negative).
    \item \textbf{FADS-Interaction (Ours)}: Extended FADS considering intersectional groups (e.g., Female-White) with $K=64$ clusters and minimum 2 samples per demographic combination.
    \item \textbf{FADS-Adaptive (Ours)}: Adaptive variant that reduces minimum sample requirement to 1 for underrepresented groups ($<5\%$).
\end{itemize}

\paragraph{Fairness Metrics.}
We measure fairness across three demographic attributes using:


\begin{itemize}[leftmargin=*]
    \item \textbf{MaxAccGap / AD (Accuracy Difference)}:
    
    \begin{center}
        $\max_a \text{Accuracy}(a) - \min_a \text{Accuracy}(a)$
    \end{center}
\end{itemize}

Lower values indicate better fairness (0 = perfect fairness, 1 = maximum unfairness).

\paragraph{Further Studies.}
We conduct ablations on: (1) \textit{dataset size} (2.5K vs 5K), (2) \textit{sensitive attribute selection} (gender/race/ethnicity), (3) \textit{number of clusters} ($K \in \{16, 32, 64, 128\}$), (4) \textit{minimum samples per combination} (1/2/5), and (5) \textit{number of shots} (0/4/8/16).

\subsection{Main Results}
\label{sec:main-results}

\noindent\textbf{Conclusion 1: Existing DS methods cannot bring consistent fairness improvement for medical image analysis.}
\medskip

Across methods, conventional DS heuristics (Random, Similarity, K-means) produce unstable and sometimes contradictory fairness outcomes on the Glaucoma benchmark (Table~\ref{tab:glaucoma_qwen} and Fig.~\ref{fig:finding1}). For example, Similarity obtains the highest task accuracy (65.3\%) but exhibits the largest Ethnicity AD (15.03\%); K-means reduces Gender and Race ADs in some cases but leaves Ethnicity AD high (12.70\%). Random selection can reduce one attribute gap while worsening another. These results indicate that optimizing for semantic relevance or cluster coverage alone does not guarantee equitable subgroup representation.

\noindent\textbf{Analysis.} The root cause is that DS heuristics optimize different objectives (relevance or coverage) without demographic constraints. Consequently, the exemplar pool chosen by a method can be balanced on one attribute but heavily skewed on another, producing attribute-specific AD variability. Therefore, fairness under ICL requires explicit demographic balancing during exemplar selection rather than relying solely on semantic heuristics.


\subsection{Scaling Analysis}
\label{sec:scaling}

\noindent\textbf{Setup.} To assess how dataset scale affects fairness and FADS performance, we compare experiments trained (or subsampled) at two dataset sizes: 2.5K and 5K samples (training pool used to draw demonstrations). We report per-attribute AD (Race, Ethnicity) at both scales and the absolute change in percentage points (pp). The result is presented in Table~\ref{tab:scaling_race}.

\noindent\textbf{Conclusion 2: Increasing dataset size generally reduces Race AD for most methods. Notably, FADS shows the largest AD changing when the training pool doubles (2.5K $\to$ 5K), indicating that fairness-aware selection benefits from larger and more diverse candidate pools.}

\begin{table*}[h]
\centering
\caption{Scaling analysis: Impact of dataset size on Race AD. Change in AD from 2.5K to 5K samples.}
\label{tab:scaling_race}
\begin{tabular}{lccc}
\toprule
Method & 2.5K Race AD (\%) & 5K Race AD (\%) & Change (pp) \\
\midrule
Zero-shot & 34.90 & 23.29 & -11.61 \\
Random & 32.49 & 21.54 & -10.95 \\
Similarity & 38.02 & 23.29 & -14.73 \\
FADS & 36.46 & 21.54 & \textbf{-14.92} \\
\bottomrule
\end{tabular}
\end{table*}

\begin{table*}[htbp]
\centering
\caption{Hyper-parameter analysis: Impact of shot budget (4-shot vs 8-shot) on accuracy and fairness metrics. All values in \%.}
\label{tab:icl_performance_ad}
\begin{tabular}{llccc}
\toprule
Method & Shots & Accuracy & Gender AD $\downarrow$ & Race AD $\downarrow$ \\
\midrule
Baseline & Zero-shot & 64.90\% & 5.14\% & 11.33\% \\
\midrule
Random & 4-shot & \textbf{64.60\%} & 3.62\% & \textbf{5.69\%} \\
Similarity & 4-shot & 64.20\% & 3.10\% & 6.97\% \\
FADS & 4-shot & \textbf{64.60\%} & \textbf{2.80\%} & 7.35\% \\
\midrule
Random & 8-shot & \textbf{65.50\%} & 4.49\% & 4.79\% \\
Similarity & 8-shot & 65.00\% & 4.14\% & 5.56\% \\
FADS & 8-shot & 65.30\% & \textbf{3.21\%} & \textbf{4.15\%} \\
\bottomrule
\end{tabular}
\end{table*}

\subsection{Hyper-parameter Analysis}
\label{sec:shots}

\noindent\textbf{Setup.} We study how the exemplar budget (number of shots) affects both performance and fairness. We evaluate 4-shot and 8-shot settings for Random, Similarity, and FADS; zero-shot is included as baseline. Results are summarized in Table~\ref{tab:icl_performance_ad}. We find

\begin{itemize}
    \item \textbf{4-shot regime:} FADS matches or slightly improves accuracy over Random/Similarity (both 64.6\% vs Zero-shot 64.9\%) while achieving the lowest Gender AD (2.80\% vs Random 3.62\%, Similarity 3.10\%). Race AD remains higher for FADS in this tight-budget setting (7.35\%), indicating a trade-off when exemplars are severely limited.
    \item \textbf{8-shot regime:} With more exemplars, FADS attains better fairness on Race (4.15\%) while preserving competitive accuracy (65.3\%). In contrast, Random and Similarity show mixed race/gender trade-offs (see Table~\ref{tab:icl_performance_ad}).
\end{itemize}

\noindent\textbf{Conclusion 3: The hyper-parameter analysis shows that FADS is robust across shot budgets: it reduces Gender AD in very low-shot regimes and reduces Race AD when more exemplars are allowed. Practically, this suggests tuning the exemplar budget depending on the primary fairness objective (e.g., prioritize gender parity at small budgets, race parity with larger budgets).}

\subsection{Data Bias Mitigation Analysis}
\label{sec:data_bias_mitigation}

\begin{table}[h]
\centering
\caption{Max Diff (\%) comparison across sensitive attributes. 
Lower values indicate smaller demographic imbalance among selected demonstrations. Here we use extended FADS that considers intersectional demographic groups (e.g., Female-White).}
\label{tab:maxdiff}
\begin{tabular}{lccc}
\toprule
\textbf{Method} & \textbf{Gender} & \textbf{Race} & \textbf{Ethnicity} \\
\midrule
Random & 11.18 & 67.30 & 86.31 \\
Similarity & 12.05 & 65.88 & 82.31 \\
K-means & \textbf{0.00} & 75.00 & \textbf{75.00} \\
FADS-Interactive & \textbf{0.00} & \textbf{0.00} & 88.22 \\
\bottomrule
\end{tabular}
\end{table}

\noindent\textbf{Conclusion 4: FADS effectively mitigates data bias in demonstration selection, enabling stable fairness improvement.}
\medskip

\noindent
To understand why FADS consistently outperforms other demonstration selection (DS) methods in fairness, we analyze the demographic balance of the selected demonstrations using the \textbf{Max Diff (\%)} metric, which quantifies the largest inter-group proportion gap among sensitive attributes (Gender, Race, Ethnicity). A lower Max Diff indicates a more balanced sample distribution within the constructed demonstration set.

\begin{figure*}[t]
  \centering
  \includegraphics[width=\linewidth]{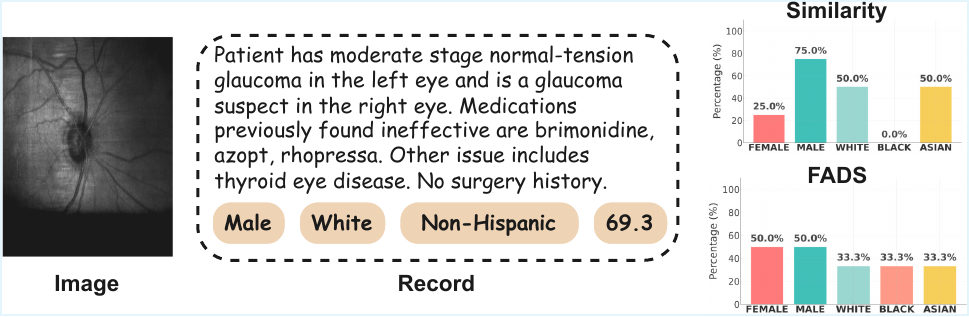}
  \caption{Case study: Similarity-based ICL vs FADS. Similarity overrepresents majority groups (75\% Male, 50\% White) and excludes the Black subgroup; FADS enforces balanced sampling (50/50 gender, 33.3\% per race).}
  \label{fig:case_study}
\end{figure*}

\noindent
As shown in Table~\ref{tab:maxdiff}, FADS selects a more demographically balanced set of exemplars compared to methods like Random, Similarity, and K-means. This is particularly evident in the \textbf{Race} dimension, where FADS outperforms other methods by reducing the racial imbalance, as seen in the significant reduction in Max Diff values. For example, Random and Similarity have high Max Diff values in Race (67.30\% and 65.88\%, respectively), whereas FADS achieves near-perfect balance, especially in the case of Race. On the other hand, K-means, despite achieving perfect gender balance by coincidence, still exhibits high racial imbalance due to clustering predominantly around the majority groups, such as White samples.

\noindent
\textbf{Analysis.} Random and Similarity sampling methods inherit the intrinsic biases of the labeled dataset, which often results in imbalanced representation across sensitive attributes. In contrast, K-means, although semantically diverse, amplifies this imbalance by clustering around dense embedding regions that are dominated by majority groups. FADS, however, integrates demographic stratification and balanced sampling within clusters, ensuring that the selected demonstrations are more demographically representative. This approach fundamentally reduces the data bias that typically underlies fairness degradation.

\noindent\textbf{Implications.} By reducing Max Diff in the selection of demonstrations, FADS directly leads to lower disparity in model outputs across demographic groups (lower AD). This demonstrates that the fairness improvement achieved by FADS is not coincidental but stems from a systematic approach to balancing demographic representation in the demonstration set. These results highlight the importance of explicitly addressing data imbalance during in-context learning, particularly in medical image analysis, to ensure fair and equitable reasoning.

\subsection{Case Study: Balanced Demonstration Selection with FADS}
\label{sec:case_study}

\noindent\textbf{Finding 4: FADS produces more demographically balanced demonstrations, directly enhancing fairness.}
\medskip

\noindent
To qualitatively illustrate how FADS mitigates data bias during in-context learning, we present a representative case from the Glaucoma dataset (Figure~\ref{fig:case_study}). The figure compares the composition of selected demonstrations under a conventional similarity-based method and under our FADS approach. The input case describes a male, White, non-Hispanic patient with moderate-stage glaucoma and comorbidities.

\noindent
This qualitative example confirms our quantitative analyses: constructing balanced and semantically relevant demonstration sets reduces downstream disparity (lower AD) and yields more interpretable conditional behavior.

To summarize, FADS consistently provides a favorable fairness--performance trade-off under realistic shot budgets and benefits further from larger candidate pools. The scaling and hyper-parameter studies show how dataset size and exemplar budget influence fairness–performance dynamics, offering practical guidance for deployment.


\section{Related Work}

\subsection{Multimodal Large Language Models}

Multimodal large language models (MLLMs) have recently emerged as powerful architectures that integrate visual, textual, and sometimes structured modalities into a unified reasoning framework, and have achieved strong performance on a wide range of vision understanding and reasoning tasks~\cite{yin2024survey,zhang-etal-2024-mm}. General-purpose models such as LLaVA, built via visual instruction tuning~\cite{liu2023visual}, show that aligning a vision encoder with an instruction-tuned LLM enables open-ended visual dialogue and chain-of-thought style reasoning. In the medical domain, early biomedical vision–language models such as BioViL and its temporal extension~\cite{boecking2022making,bannur2023learning} and GLoRIA~\cite{huang2021gloria} learn joint representations between radiology images and reports, while more recent generalist medical MLLMs, including LLaVA-Med~\cite{li2023llava} and Hulu-Med~\cite{jiang2025hulumedtransparentgeneralistmodel}, adapt this paradigm to diverse clinical images and instructions and have shown significant improvements in zero-shot reasoning and robustness across imaging modalities including fundus, histopathology, and CT scans. Proprietary systems such as Med-PaLM and Med-PaLM~2~\cite{singhal2023large,singhal2025toward} further demonstrate near-expert performance on medical question answering through instruction tuning on large-scale expert-curated clinical data.

Despite their remarkable progress, the trustworthiness of medical MLLMs, such as fairness, robustness and hallucination, remains a major open challenge~\cite{HE2025102963,huang2024trustllm, xia2024cares, wu2024fmbench}. EchoBench~\cite{yuan2025echobench} further reveals severe sycophancy in medical MLLMs, where models tend to agree with user-provided misinformation. 
However, mitigating these trustworthiness issues is non-trivial in practice. Most foundation-scale MLLMs operate as black boxes, hindering transparency and interpretability in clinical decision-making and limiting the possibility of model-specific interventions. Moreover, they remain highly sensitive to domain shift between natural and medical imagery, often exhibiting degraded robustness and fairness across institutions, imaging devices, or demographic subgroups~\cite{musa2025addressing,yang2025demographic}. Finally, adapting MLLMs through fine-tuning or domain-specific retraining is computationally expensive and risks catastrophic forgetting, resulting in erasing general multimodal reasoning skills in favor of narrow task specialization~\cite{zhai2023investigating,huang2024learn,wu2025mitigating}. These challenges motivate the need for tuning-free, data-efficient mechanisms, such as our proposed fairness-aware in-context learning framework, that can adapt existing MLLMs for equitable medical reasoning without compromising their generalization capacity or requiring access to model parameters.

\subsection{Fairness in Medical Image Analysis}

Fairness in medical image analysis has become an increasingly critical research focus, as biased AI systems risk amplifying existing health disparities and eroding clinical trust. Systematic investigations~\cite{drukker2023toward,koccak2025bias} have revealed that demographic, institutional, and device-level biases persist throughout the medical AI pipeline—from data collection to model deployment. These biases often manifest as uneven predictive accuracy and sensitivity across subpopulations defined by gender, race, ethnicity, or socioeconomic status. Empirical analyses further attribute such disparities to dataset imbalance, missing demographic annotations, and domain-specific artifacts such as scanner heterogeneity or site-dependent acquisition protocols. As a result, fairness violations can arise even when models achieve high overall accuracy, underscoring the limitations of standard evaluation metrics in safety-critical domains.

To address these inequities, a variety of fairness-enhancing strategies have been proposed. Data-centric approaches employ reweighting, resampling, or augmentation to balance demographic representation, while model-centric methods leverage adversarial debiasing or gradient reversal to decorrelate latent features from sensitive attributes. Domain adaptation and generalization frameworks attempt to align feature distributions across institutions, and federated learning schemes such as \citet{zhang2024unified} enable decentralized training while preserving population diversity. More recent efforts, including \citet{deng2023fairness} and \textit{FairCLIP}~\cite{luo2024fairclip}, explore representation-level fairness through multimodal contrastive learning, showing that bias can persist even in pretrained embeddings. However, most existing methods are computationally intensive, require large-scale demographic annotations, or depend on model retraining—making them impractical for closed-source or foundation-level MLLMs. Furthermore, these approaches often trade off fairness for accuracy or interpretability. In contrast, our proposed \textbf{Fairness-Aware Demonstration Selection (FADS)} framework adopts a data- and model-agnostic perspective, leveraging in-context learning to achieve fairness improvements without fine-tuning. This paradigm offers a lightweight, scalable alternative that complements ongoing efforts toward trustworthy and equitable medical AI.


\section{Discussion and Conclusion}
\label{sec:discussion}

In this work, we investigated the potential of \textit{In-Context Learning (ICL)} to enhance fairness in multimodal large language models (MLLMs) for medical image reasoning. Through systematic analysis, we revealed that existing demonstration selection strategies—such as random, similarity-based, and clustering-based methods—often yield inconsistent fairness outcomes due to demographic imbalances in the selected exemplars. To address this challenge, we proposed the \textbf{Fairness-Aware Demonstration Selection (FADS)} framework, which constructs balanced and semantically relevant demonstration sets by integrating demographic-aware clustering with similarity-based sampling. FADS improves fairness without model retraining or additional data collection, offering a lightweight, tuning-free, and scalable approach suitable for foundation-level MLLMs. These results highlight the potential of fairness-aware in-context learning as a scalable and data-efficient solution for equitable medical image reasoning.

Comprehensive experiments across multiple medical imaging benchmarks demonstrate that FADS consistently reduces gender-, race-, and ethnicity-related disparities while maintaining competitive task performance. Our analyses further show that balancing demographic composition at the demonstration level directly mitigates both data-driven and model-induced biases, enabling more equitable in-context reasoning. These findings highlight the promise of fairness-aware ICL as a practical paradigm for promoting trustworthy and inclusive medical AI. In future work, we plan to extend FADS to handle multiple sensitive attributes and more complex intersectional fairness settings. Another promising direction is to explore adaptive demonstration retrieval strategies that dynamically balance fairness and performance at inference time. We believe such advances will further strengthen fairness-aware in-context learning as a general framework for equitable and transparent medical AI.


{
    \small
    \bibliographystyle{ieeenat_fullname}
    \bibliography{main}
}


\end{document}